\documentclass[a4paper, conference, 10pt]{IEEEtran}
\pdfoutput=1
\usepackage{cite}
\usepackage[pdftex]{graphicx}
\graphicspath{{./images/}}
\DeclareGraphicsExtensions{.pdf,.jpeg,.png,.jpg}
\usepackage[cmex10]{amsmath}
\usepackage{cite}
\usepackage{url}
\usepackage{multirow}
\usepackage{subfigure}
\usepackage{amsmath,amssymb}
\newtheorem{thm}{Lemma}

\newtheorem{prop}{Proposition}

\usepackage{times}

\begin{document}
\title{Risk Estimation Without Using Stein's Lemma -- Application to Image Denoising}
\author{\IEEEauthorblockN{Sagar~Venkatesh~Gubbi$^1$ and Chandra Sekhar Seelamantula$^2$}
\IEEEauthorblockA{$^1$Department of Electrical Communication Engineering\\
$^2$Department of Electrical Engineering\\
Indian Institute of Science, Bangalore - 560 012, India\\
Email: sagar@ece.iisc.ernet.in, chandra.sekhar@ieee.org}}

%\markboth{}%{Gubbi and Seelamantula: Risk Estimation Without Using Stein's Lemma -- Application to Image Denoising}

%\IEEEoverridecommandlockouts 
%\IEEEpubid{\makebox[\columnwidth]{978-1-4799-6619-6/15/\$31.00~\copyright~2015 IEEE \hfill}
%\hspace{\columnsep}\makebox[\columnwidth]{ }}
\maketitle

\begin{abstract}
We address the problem of image denoising in additive white noise without placing restrictive assumptions on its statistical distribution. 
In the recent literature, specific noise distributions have  been considered and correspondingly, optimal denoising techniques have been developed. One of the successful approaches for denoising relies on the notion of unbiased risk estimation, which enables one to obtain a useful substitute for the mean-square error. For the case of additive white Gaussian noise contamination, the risk estimation procedure relies on Stein's lemma. Sophisticated wavelet-based denoising techniques, which are essentially nonlinear, have been developed with the help of the lemma. We show that, for linear, shift-invariant denoisers, it is possible to obtain unbiased risk estimates of the mean-square error without using Stein's lemma. An interesting consequence of this development is that the unbiased risk estimator becomes agnostic to the statistical distribution of the noise. As a proof of principle, we show how the new methodology can be used to optimize the parameters of a simple Gaussian smoother. By locally adapting the parameters of the Gaussian smoother, we obtain a shift-variant smoother, which has a denoising performance (quantified by the improvement in peak signal-to-noise ratio (PSNR)) that is competitive to far more sophisticated methods reported in the literature. The proposed solution exhibits considerable parallelism, which we exploit in a Graphics Processing Unit (GPU) implementation.
\end{abstract}

\IEEEpeerreviewmaketitle

\section{Introduction}
\indent The need for denoising images is frequently encountered in applications such as low-light photography, optical imaging, microscopy applications, biomedical imaging modalities such as magnetic resonance imaging and ultrasound imaging, synthetic aperture radar imaging, astronomical imaging, etc. The nature of noise distortion and its statistical properties are determined by the physics of image acquisition. While the additive white Gaussian noise model is frequently considered, there are many cases in which the assumption does not hold, for example, the noise is multiplicative and exponential or gamma distributed in coherent imaging \cite{bioucasfigueredo}, the noise statistics follow a Poisson distribution in low-photon-count microscopy applications and is neither additive nor multiplicative \cite{purelet}, the noise in magnetic resonance imaging follows a chi-square distribution \cite{cure}, etc. Noise not only affects the visual quality of an image, but also distorts the features that one computes for subsequent tasks related to image analytics or pattern classification in the application chain. The goal in image denoising is to suppress noise and smooth the image by minimally affecting edges and texture present in the image.\\
\indent In this paper, we consider the additive noise model with the noise being white and possessing zero mean and finite variance. We do not place any distributional assumptions on the noise. Before proceeding with the developments, we review some recent literature on the topic.

\subsection{Prior Art}
\indent Wavelet thresholding has been the most popular transform-domain denoising approach. Donoho and Johnstone proposed VisuShrink \cite{visushrink}, which is a point-wise thresholding technique based on a universal threshold that is a function of the noise variance and the number of pixels. They also proposed SUREShrink \cite{sureshrink}, in which they derived an optimal threshold that minimizes Stein's Unbiased Risk Estimation (SURE), while VisuShrink minimizes the minimax error. Chang et al. proposed BayesShrink \cite{bayesshrink}, wherein they modeled the wavelet coefficients as a realization of a generalized Gaussian distribution and derived a subband-adaptive soft-thresholding function. Portilla et al. modeled the wavelet coefficients at adjacent positions and scales as a Gaussian scale mixture (GSM) and used a Bayesian least-squares (BLS) cost \cite{portilla}. Pi\v{z}urica and Philips proposed ProbShrink \cite{probshrink} wherein they modeled the noise-free signal by a generalized Laplacian distribution and derived a threshold based on the estimated probability that a given wavelet coefficient contains significant information. Sendur and Selesnick proposed BiShrink \cite{bishrink1,bishrink2} where they modeled interscale dependencies using non-Gaussian bivariate distributions. Starck et al. \cite{starck} proposed curvelet transforms, which showed superior denoising performance at the cost of higher memory and processing time. Luisier et al. proposed a SURE-optimal pointwise thresholding technique in the orthonormal wavelet domain taking into account interscale wavelet dependencies \cite{luisier}. They also proposed the idea of {\it linear expansion of thresholds} (LET) \cite{surelet}, a novel approach for designing nonlinear thresholding functions in a computationally efficient manner. Zhang and G\"unturk \cite{zhang} proposed a multiresolution bilateral filter (MBF) method, a hybrid approach in which they perform thresholding of the wavelet coefficients (the detail subbands) and bilateral filtering of the approximation coefficients. Dabov et al. proposed BM3D \cite{bm3d}, which is a patch-based framework to perform block matching and 3D collaborative filtering in the transform domain. It is among the best performing techniques in the state of the art. Kishan and Seelamantula \cite{harini} optimized the parameters of the bilateral filter using the SURE criterion. Elad and Aharon proposed a dictionary-based approach to perform image denoising \cite{elad}. Chatterjee and Milanfar presented a clustering-based framework for denoising \cite{milanfar1} and proposed a lower bound on the mean-square error (MSE) at the output of a denoising function \cite{milanfar2}. They also developed patch-based Wiener filters for performing near-optimal image denoising \cite{milanfar3}.

\subsection{Contributions}
\indent We consider the ground-truth or the clean image to be deterministic and the noise to be random (Section~\ref{probform}). We develop an estimate of the MSE first using Stein's lemma \cite{stein}, which relies on Gaussian statistics for noise (Section~\ref{msestein}). We then develop a {\it Stein-free} MSE estimation approach, which does not impose any distribution on noise (Section~\ref{msesteinfree}). We show that the estimated MSE closely follows the oracle MSE. To illustrate the applicability of the result, and to serve as proof of principle, we consider optimizing the parameters of a simple Gaussian smoother on a patch-by-patch basis, resulting in an overall spatially-varying Gaussian smoother (Section~\ref{svg}). The denoising performance turns out to be competitive with the state of the art. We show results considering both Gaussian and Laplacian noise cases, and demonstrate that the change in noise distribution does not alter the denoising capability of the proposed approach, whereas risk estimation approaches such as the SURE-LET show deterioration in performance (Section~\ref{res}). The proposed denoising approach also allows for parallelization using Graphics Processing Units (GPUs) \cite{pharr2005gpu}, which we have also used in the implementation.

\section{Problem Formulation}
\label{probform}
\indent Consider an $N$-dimensional noise-free deterministic image ${\bf x}$ (we consider column-vectorized representation of an image) corrupted by additive noise ${\bf w}$, the entries of which are assumed to have zero mean, finite variance $\sigma^2$, and mutually uncorrelated. The noisy image is ${\bf y} = {\bf x} +{\bf w}$. The denoising function is, in general, represented by ${\bf f}$ and the denoised output is ${\hat{\bf x}} = {\bf f}({\bf y})$. The denoising operator can be nonlinear in general, but in this paper we are interested in linear, shift-invariant denoising functions, that is, $\hat{\bf x} = {\bf f}({\bf y}) = \mathbb{H}{\bf y}$, where $\mathbb{H}$ is a Toeplitz matrix constructed from a stable filter ${\bf h}$. Note that ${\bf f}$ is actually a collection of $N$ denoising functions with the $k^{th}$ element expressed as $f_k({\bf y})$. The goal is to optimize the parameters of the filter (in the simplest case, its impulse response) such that the MSE between ${{\bf x}}$ and ${\hat{\bf x}}$ is minimized. The MSE is $\mathcal{C} = \mathcal{E} \{ \|{\bf x}  - \widehat{\bf x} \|^2 \}.$ Developing the squares, we obtain 
\begin{equation}
\mathcal{C} = \| {\bf x}\|^2 + \mathcal{E} \{\|\widehat{\bf x} \|^2 \} - 2\mathcal{E} \{ {\bf x}^{\text {\sc t}}\widehat{\bf x}\}.
\end{equation}
Since ${\bf x}$ is deterministic, $\| {\bf x}\|^2$ simply adds a bias to the cost and does not affect minimization of $\mathcal{C}$ with respect to ${\bf h}$. The term $\mathcal{E} \{\|\widehat{\bf x} \|^2 \}$ can be estimated for a chosen ${\bf h}$. The term $\mathcal{E} \{ {\bf x}^{\text {\sc t}}\widehat{\bf x}\}$ cannot be computed since ${\bf x}$ is not known; at best, it can be estimated and we would like to obtain an unbiased estimate.

\section{Estimating the MSE}
\label{mseest}
\subsection{The Stein Approach}
\label{msestein} 
\indent The Stein approach requires a Gaussian assumption on ${\bf w}$ and makes use of the following lemma from \cite{stein}. 
\begin{thm} (Stein, 1981)
Let $Y$ be a ${\cal N}(0,\sigma^2)$ real random variable and let $f: \mathbb{R}\rightarrow\mathbb{R}$ be an indefinite integral of the Lebesgue measurable function $f'$, essentially the derivative of $f$. Suppose also that $\mathcal{E}\{|f'(Y)|\}<\infty$. Then
\begin{equation}
\mathcal{E}\{Yf(Y)\}=\sigma^2 \mathcal{E}\{f'(Y)\}.   \nonumber
\label{epsilon}
\end{equation}
\end{thm}
The proof of the lemma is a direct consequence of an identity satisfied by the Gaussian density followed by application of the {\it integration-by-parts} formula. Stein's lemma facilitates estimation of the mean of $Yf(Y)$ in terms of the mean of $f'(Y)$.\\
\indent For vectors (or vectorized images), we make use of the multidimensional version of the Stein lemma, which was provided by Luisier et al. \cite{luisier}:
\begin{equation}
\mathcal{E}\{{\bf x}^{\text{\sc t}}{\bf f}({\bf y})\} = \mathcal{E}\left\{{\bf y}^{\text{\sc t}}{\bf f({\bf y})} - \sigma^2\, \sum_{k} \displaystyle\frac{\partial f_k}{\partial y_k}\right\} %\mbox{div}_{\bf y}\left({\bf f}({\bf y})\right)\},
\end{equation}
assuming that $\mathcal{E}\left\{\displaystyle\frac{\partial f_{k}({\bf y})}{\partial {\bf y}_k}\right\} < \infty, \forall k$, where $y_k$ denotes the $k^{th}$ element of ${\bf y}$.
The multidimensional version allows one to write Stein's unbiased risk estimate (SURE) for the MSE as follows:
\begin{eqnarray}
\mbox{SURE} &=& \|\mathbb{H}{\bf y}-{\bf y}\|^2 + 2\sigma^2\, \sum_{k} \displaystyle\frac{\partial f_k}{\partial y_k} - N\sigma^2.
\nonumber
\end{eqnarray}
The value of $f_k$ is simply the inner product between the $k^{th}$ row of $\mathbb{H}$ and ${\bf y}$. Hence, $\displaystyle \sum_k\frac{\partial f_k}{\partial y_k} = Nh_0$, where $h_0$ is the entry of ${\bf h}$ at index 0 (the first entry of ${\bf h}$). Hence, SURE in this case becomes
\begin{eqnarray}
\mbox{SURE} &=& \|\mathbb{H}{\bf y}-{\bf y}\|^2 + 2\sigma^2\,N\,h_0 - N\sigma^2.
\label{sureexp}
\end{eqnarray}
SURE is an unbiased estimate of the MSE and its variance is small since there are a large number of pixels in practical images. Hence, SURE is a reliable surrogate for the MSE. Note that Stein's lemma is based on the assumption of Gaussian noise statistics. For other noise types, one must derive corresponding risk estimators --- this has been an active area of research over the past decade \cite{luisier,surelet,cure,mure1,mure2}.

\subsection{The Stein-Free Approach}
\label{msesteinfree}
\begin{figure}[!t]
  \centering
  \includegraphics[width=0.8\linewidth]{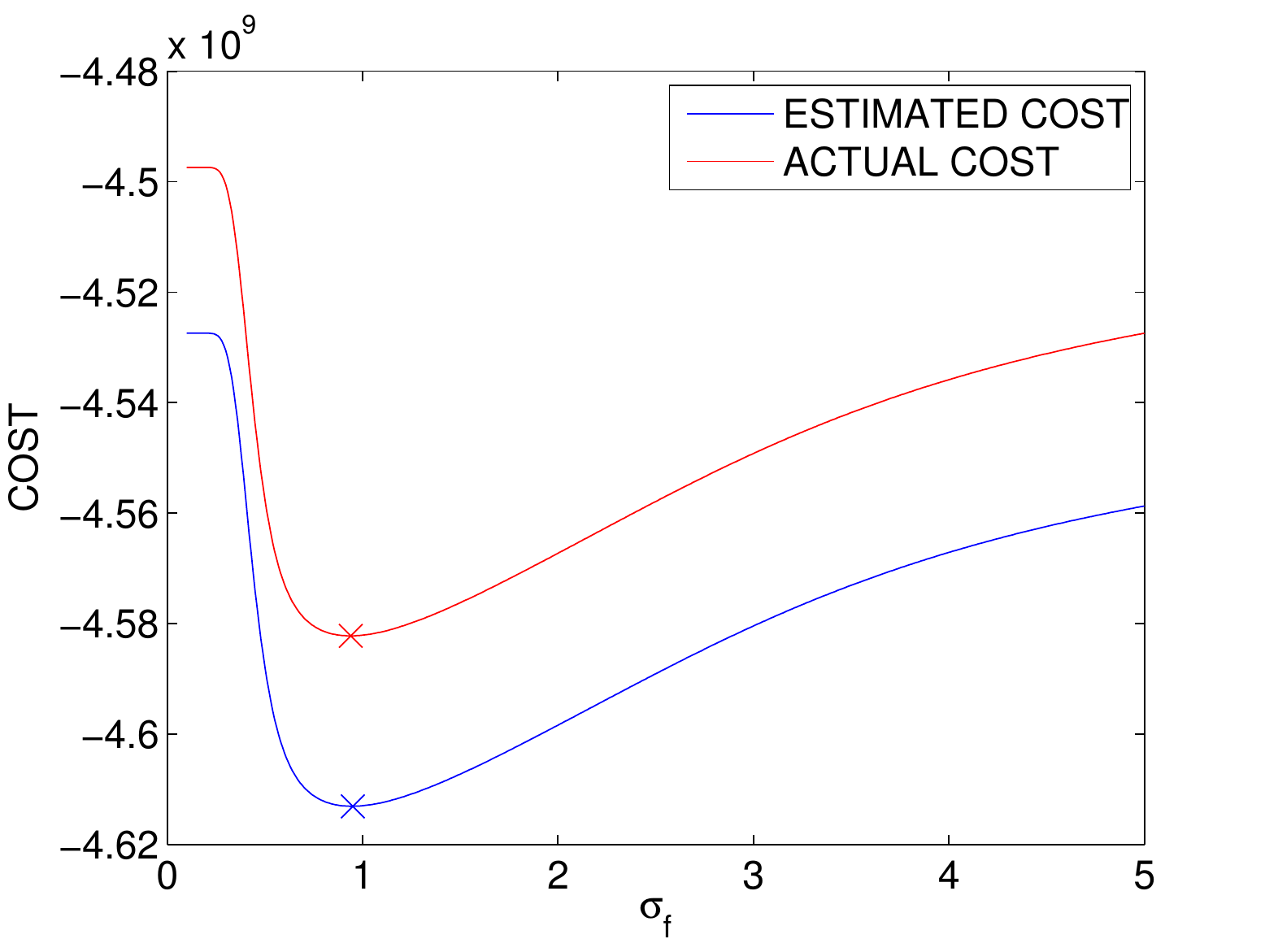}
  \caption{The Stein-free unbiased estimate of the cost and the true cost for \textit{Lenna} 512$\times$512 image corrupted by additive white Gaussian noise (AWGN) of standard deviation 20. A bias is added to the actual cost to aid visualization. The term $\|{\bf x}\|^2$ has been suppressed, since it has no effect on minimization (also the reason why the y-axis is negative).}
  \label{costPlot}
\end{figure}

\indent Consider the term $\mathcal{E}\{{\bf x}^{\text{\sc t}}{\bf f}({\bf y})\}$ and substituting ${\bf x} = {\bf y} - {\bf w}$, and ${\bf f}({\bf y}) = \mathbb{H}{\bf y}$, we get
\begin{eqnarray}
\mathcal{E}\{{\bf x}^{\text{\sc t}}{\bf f}({\bf y})\} %&=& \mathcal{E}\{{\bf y}^{\text{\sc t}}{\bf f}({\bf y})\} - \mathcal{E}\{{\bf w}^{\text{\sc t}}{\bf f}({\bf y})\}\nonumber\\
&=& \mathcal{E}\{{\bf y}^{\text{\sc t}}\mathbb{H}{\bf y}\} - \mathcal{E}\{{\bf w}^{\text{\sc t}}\mathbb{H}{\bf y}\}.
\end{eqnarray}
Next, considering $\mathcal{E}\{{\bf w}^{\text{\sc t}}\mathbb{H}{\bf y}\}$ and substituting ${\bf y} = {\bf x} + {\bf w}$, we get
\begin{eqnarray}
\mathcal{E}\{{\bf w}^{\text{\sc t}}\mathbb{H}{\bf y}\} &=& \mathcal{E}\{{\bf w}^{\text{\sc t}}\mathbb{H}{\bf x}\} + \mathcal{E}\{{\bf w}^{\text{\sc t}}\mathbb{H}{\bf w}\}\nonumber\\
&=& N\,\sigma^2\,h_0,
\end{eqnarray}
where we have used the assumptions that ${\bf x}$ is deterministic and that ${\bf w}$ has zero-mean and uncorrelated entries. Hence, we have 
\begin{eqnarray}
\mathcal{E}\{{\bf x}^{\text{\sc t}}{\bf f}({\bf y})\} %&=& \mathcal{E}\{{\bf y}^{\text{\sc t}}{\bf f}({\bf y})\} - \mathcal{E}\{{\bf w}^{\text{\sc t}}{\bf f}({\bf y})\}\nonumber\\
&=& \mathcal{E}\{{\bf y}^{\text{\sc t}}\mathbb{H}{\bf y}\} - N\,\sigma^2\,h_0,
\end{eqnarray}
indicating that ${\bf y}^{\text{\sc t}}\mathbb{H}{\bf y} - N\,\sigma^2\,h_0$ is an unbiased estimator of $\mathcal{E}\{{\bf x}^{\text{\sc t}}{\bf f}({\bf y})\}$. Also, since ${\bf y} = {\bf x + w}$, $\|{\bf x}\|^2 = \mathcal{E}\{\|{\bf y}\|^2\} - N\sigma^2$.
Putting the pieces together, we have
\begin{eqnarray}
\mathcal{C} &=& \mathcal{E}\{\|{\bf y}\|^2\} - N\sigma^2 + \mathcal{E} \{\|\mathbb{H}{\bf y} \|^2 \}\nonumber\\
&& - 2\mathcal{E}\{{\bf y}^{\text{\sc t}}\mathbb{H}{\bf y}\} + 2 N\,\sigma^2\,h_0,\nonumber\\
&=& \mathcal{E}\{\|\mathbb{H}{\bf y} - {\bf y}\|^2\} + 2\sigma^2\,N\,h_0 - N\sigma^2,
\end{eqnarray}
which allows us to infer that $\hat{\mathcal{C}} = \|\mathbb{H}{\bf y} - {\bf y}\|^2 + 2\sigma^2\,N\,h_0 - N\sigma^2$ is an unbiased estimator of $\mathcal{C}$, which is also computable in the sense that it depends only on the observed image ${\bf y}$, the filter ${\bf h}$ (or equivalently $\mathbb{H}$) and noise variance $\sigma^2$. We observe that $\hat{\mathcal{C}}$ is identical to SURE. However, the SURE was derived under the Gaussian noise assumption, whereas we did not make any such assumption in our derivation. We have only assumed knowledge of the first- and second-order statistics of noise.\\
\indent We summarize the result in the form of the following proposition.
\begin{prop}
Let ${\bf y = x+w}$, where ${\bf x}$ is an $N$-dimensional deterministic vector and ${\bf w}$ is an $N$-dimensional random vector with uncorrelated entries possessing zero mean and finite variance $\sigma^2$. Let $\mathbb{H}$ denote a stable linear Toeplitz operator such that $\mathcal{E}\{\|\mathbb{H}{\bf y}\|^2\}$ is finite. Then, $\|\mathbb{H}{\bf y} - {\bf y}\|^2 + 2\sigma^2\,N\,h_0 - N\sigma^2$ is an unbiased estimate of $\mathcal{C} \stackrel{\Delta}= \mathcal{E}\{\|{\bf x - \mathbb{H}{\bf y}}\|^2\}$, where $h_0$ denotes the principal diagonal entry of $\mathbb{H}$.
\end{prop}
\indent In order to illustrate that minimizing $\hat{\mathcal{C}}$ is nearly equivalent to minimizing $\mathcal{C}$, we consider the {\it Lenna} image of size 512 $\times$ 512 pixels, corrupted by additive white Gaussian noise of standard deviation 20. Considering 
$\mathbb{H}$ to be an isotropic Gaussian smoother with parameter $\sigma_f$, we show in Figure~\ref{costPlot} how the oracle MSE (legend: actual cost) and its Stein-free unbiased risk estimate (legend: estimated cost) vary as a function of $\sigma_f$. It is of interest to compute the optimum $\sigma_f$. The absolute difference between the optimal $\sigma_{f}$ obtained from the true cost and that obtained from its estimate was found to be less than 0.01. Thus, the proposed risk estimator is an accurate substitute for the MSE. Although we have considered Gaussian noise for illustration, in the experimental results, we shall also consider Laplacian noise contamination.

\subsection{Noise Variance Estimation}
\label{noiseest}
\indent In the theoretical developments, we have assumed prior knowledge of the noise variance $\sigma^2$. In practice, one may not know $\sigma^2$ accurately, and hence it must be estimated. Noise can be estimated from the highpass subbands. Highpass filtering suppresses smooth portions of the image, enhances the noise and edges, but it also converts the white noise to pink noise. For additive white Gaussian noise in the image domain, the standard deviation is estimated as \cite{mallat1998wavelet}
\begin{equation}
    \label{estnoise}
    \widehat{\sigma} = \frac{\mbox{median}\left\{ |{\bf h} \ast {\bf y}| \right\}}{0.6745},
\end{equation}
where ${\bf h}$ is a 2-D highpass filter specified by the mask
$$
	{\bf h} = \frac{1}{9}\left( \begin{array}{rrr}
    -1 & -1 & -1 \\
    -1 &  8 & -1 \\
    -1 & -1 & -1 \\
  	\end{array} \right).
$$
The convolution may be implemented directly in 2D without requiring vectorization. For Laplacian noise, the estimate of the standard deviation is given as \cite{laplaceest}: 
\begin{equation}
    \label{estnoise}
    \widehat{\sigma} = \frac{\mbox{median}\left\{ |{\bf h} \ast {\bf y}| \right\}}{0.4901}.
\end{equation}
The median effectively suppresses outliers in the highpass subband, which actually correspond to the edges in the image.\\
\indent In Tables~\ref{noise_estim_results1} and \ref{noise_estim_results2}, we show the noise variance estimates for white Gaussian noise and white Laplacian noise, respectively, obtained by averaging over 20 realizations of the noisy image, for various images. We observe that the estimates are sufficiently accurate to be practically useful in denoising applications.

\begin{table}[!h]
	\centering
	\caption{Estimation of noise standard deviation -- white Gaussian noise}
    \begin{tabular}{lccc}
        \hline
  %      ~                         &                      & $\widehat{\sigma}$   &                 \\ 
        Image                     & $\sigma$ = 5         & $\sigma$ = 20        & $\sigma$ = 50   \\\hline
\textit{Barbara} (512$\times$512)   & 8.17          & 22.05         & 49.11   \\ 
\textit{Boats} (512$\times$512)     & 7.47           & 20.52         & 48.23   \\ 
\textit{Cameraman} (256$\times$256) & 7.26           & 21.78         & 49.38   \\ 
\textit{Lenna} (512$\times$512)      & 6.18           & 19.78         & 47.71   \\
        \hline
    \end{tabular}
    \label{noise_estim_results1}
\end{table}

\begin{table}[!h]
	\centering
	\caption{Estimation of noise standard deviation -- white Laplacian noise}
    \begin{tabular}{lccc}
        \hline
%        ~                         &                      & $\widehat{\sigma}$   &                 \\ 
        Image                     & $\sigma$ = 5         & $\sigma$ = 20        & $\sigma$ = 50   \\\hline
\textit{Barbara} (512$\times$512)   & 10.56         & 26.19        & 55.99  \\ 
\textit{Boats} (512$\times$512)     & 9.76        & 23.88        & 54.06  \\ 
\textit{Cameraman} (256$\times$256) & 9.07         & 25.46        & 56.18  \\ 
\textit{Lenna} (512$\times$512)      & 7.72         & 22.49        & 52.91  \\
        \hline
    \end{tabular}
    \label{noise_estim_results2}
\end{table}

\section{Spatially-Varying Gaussian Smoother (SVGS)}
\label{svg}
\begin{figure}[!t]
  \centering
  \includegraphics[width=0.35\linewidth]{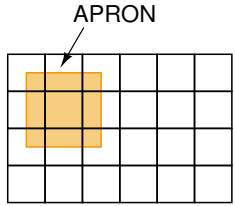}
  \caption{An illustration of a rectangular image divided into 24 blocks (6$\times$4). An ``apron" of additional pixels around the block is considered when determining filter parameters for that block.}
  \label{apron}
\end{figure}

\indent For the purpose of illustration, we consider optimizing the parameter of a Gaussian smoother based on the Stein-free risk estimator. The optimal parameter can be computed for the whole image, but we prefer to optimize it locally on a patch-by-patch basis so that it can adapt to the local structure of the image and does not smooth too much across edges. Consider the truncated Gaussian kernel
$$
h(x,y) = \frac{e^{-\frac{1}{2}\left(\frac{x_\theta^2}{\sigma_x^2} + \frac{y_\theta^2}{\sigma_y^2} \right)}}{\sum\limits_{\overline{x}=-\frac{M}{2}}^{\frac{M}{2}}\sum\limits_{\overline{y}=-\frac{M}{2}}^{\frac{M}{2}}e^{-\frac{1}{2}\left(\frac{\overline{x}_\theta^2}{\sigma_x^2} + \frac{\overline{y}_\theta^2}{\sigma_y^2} \right)}},
$$
for $-\frac{M}{2} \le x, y \le \frac{M}{2}$, and $0$ elsewhere. Here, $\theta$ is the orientation of the noisy image patch obtained from the gradient-based structure tensor approach, and
$$
x_\theta = \quad x\, \mbox{cos}\theta + y\,\mbox{sin}\theta, \quad \mbox{and}\quad y_\theta = -x\, \mbox{sin}\theta + y\,\mbox{cos}\theta.
$$
We divide the image into small blocks or patches and optimize the Gaussian filter parameter for each patch using the Stein-free approach. Consequently, within a patch, the filter remains shift-invariant and across patches, the filtering becomes shift-varying. Since the optimization can be performed for each block independently of the others, the overall optimization can be carried out in parallel. In dividing the image into small blocks, the question of choosing the block size arises. If the number of pixels per block is small, the mean-square error estimate becomes unreliable. To address this issue, we consider an ``apron" of additional pixels around each block when computing the mean-squared error estimate (cf. Figure~\ref{apron} for illustration). However, this comes at the price of slightly increased computation. We experimentally assess the time taken for denoising and peak signal-to-noise ratio (PSNR) for different block and apron sizes and correspondingly determine suitable values.

\section{Experimental Results}
\label{res}

\begin{table*}[!t]
\centering 
\caption{Comparison of denoising performance of SVGS with state-of-the-art methods (Gaussian noise)}
\begin{tabular}{cccccccc} 
\hline
Image     & Input     & \multicolumn{5}{c}{Output PSNR(dB) }  \\
 	      & PSNR(dB)  & SVGS     & MBF\cite{zhang}  & OWT\cite{luisier}  & BiShrink\cite{bishrink1} & BM3D\cite{bm3d} & ProbShrink\cite{probshrink}\\ 
\hline
\multirow{4}{*}{\includegraphics[width=0.06\linewidth]{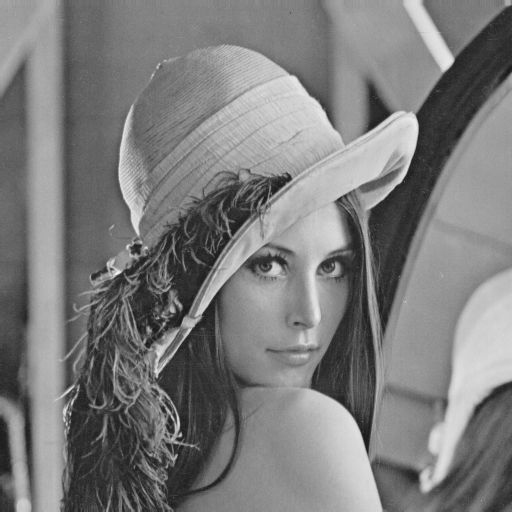}}
          & 28.13     & 34.90    & 34.47      & 34.33          & 34.31            & 35.83           & 34.92 \\ \cline{2-8}
          & 22.13     & 31.69    & 31.29      & 31.35          & 31.17            & 33.01           & 31.92 \\ \cline{2-8}
          & 14.61     & 26.84    & 26.63      & 27.16          & 27.00            & 28.77           & 27.56 \\ \cline{2-8}
          & 10.15     & 22.41    & 22.27      & 20.81          & 22.64            & 23.95           & 22.89 \\ \hline
          
\multirow{4}{*}{\includegraphics[width=0.06\linewidth]{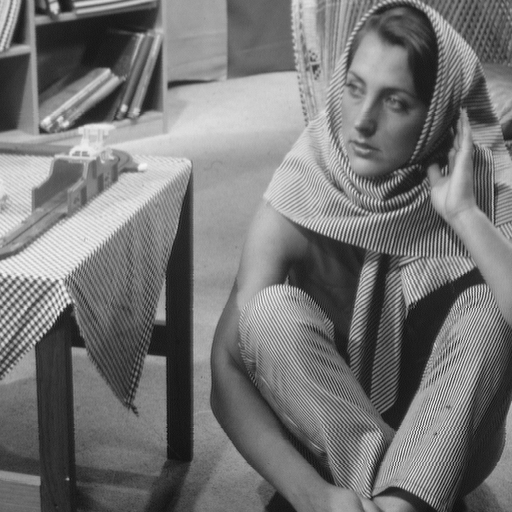}}
          & 28.13     & 32.33    & 31.66      & 29.90          & 32.13            & 34.65           & 32.35 \\ \cline{2-8}
          & 22.17     & 28.66    & 27.67      & 27.34          & 28.24            & 31.67           & 29.29 \\ \cline{2-8}
          & 14.76     & 23.92    & 23.44      & 23.54          & 23.81            & 26.92           & 24.34 \\ \cline{2-8}
          & 10.24     & 20.25    & 20.27      & 19.25          & 20.37            & 21.87           & 20.43 \\ \hline
          
\multirow{4}{*}{\includegraphics[width=0.06\linewidth]{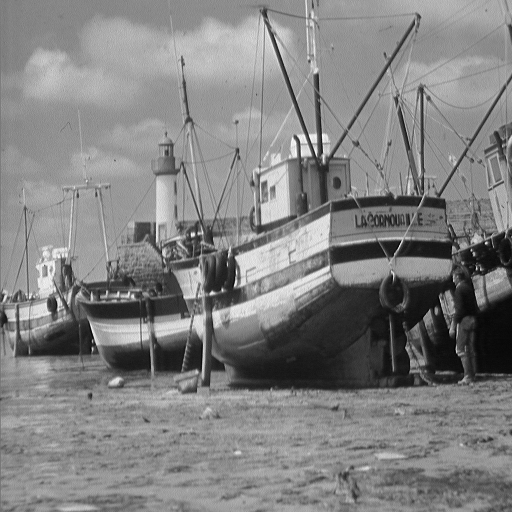}}
          & 28.14     & 32.54    & 32.52      & 31.96          & 32.36            & 33.58           & 32.85 \\ \cline{2-8}
          & 22.18     & 29.72    & 29.38      & 29.41          & 29.17            & 30.80           & 29.91 \\ \cline{2-8}
          & 14.60     & 25.01    & 24.98      & 25.31          & 24.98            & 26.46           & 25.60 \\ \cline{2-8}
          & 10.11     & 21.35    & 21.32      & 19.99          & 21.43            & 22.49           & 21.72 \\ \hline
          
\multirow{4}{*}{\includegraphics[width=0.06\linewidth]{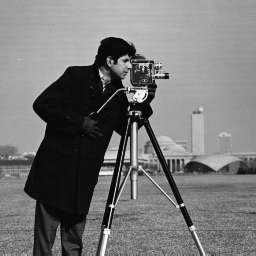}}
          & 28.27     & 32.38    & 32.61      & 30.82          & 32.40            & 33.39           & 32.06 \\ \cline{2-8}
          & 22.44     & 28.68    & 29.17      & 28.19          & 28.48            & 30.18           & 28.80 \\ \cline{2-8}
          & 14.88     & 23.30    & 23.71      & 23.52          & 23.39            & 25.21           & 23.87  \\ \cline{2-8}
          & 10.27     & 19.29    & 19.37      & 19.01          & 19.29            & 20.56           & 19.51 \\ \hline
          
\end{tabular}
\label{compare}
\end{table*}

\begin{table*}[!t]
\centering 
\caption{Comparison of SVGS with state-of-the-art methods (Laplacian noise)}
\begin{tabular}{cccccccc} 
\hline
Image     & Input     & \multicolumn{5}{c}{Output PSNR(dB) }  \\
 	      & PSNR(dB)  & SVGS     & MBF\cite{zhang}  & OWT\cite{luisier}  & BiShrink\cite{bishrink1} & BM3D\cite{bm3d} & ProbShrink\cite{probshrink}\\ 
	      \hline
\multirow{4}{*}{\includegraphics[width=0.06\linewidth]{lena_orig}}
          & 28.14     & 34.58    & 34.29      & 31.05          & 34.20            & 35.52           & 34.71 \\ \cline{2-8}
          & 22.21     & 31.79    & 30.97      & 28.71          & 30.91            & 32.91           & 30.66 \\ \cline{2-8}
          & 15.08     & 27.34    & 26.67      & 22.55          & 26.98            & 28.84           & 25.82 \\ \cline{2-8}
          & 11.04     & 23.37    & 23.18      & 13.69          & 23.55            & 24.81           & 23.94 \\ \hline
          
\multirow{4}{*}{\includegraphics[width=0.06\linewidth]{barbara_orig}}
          & 28.15     & 31.54    & 31.67      & 25.42          & 32.22            & 33.95           & 32.39 \\ \cline{2-8}
          & 22.28     & 28.20    & 27.66      & 23.48          & 28.31            & 31.27           & 29.36 \\ \cline{2-8}
          & 15.21     & 23.80    & 23.54      & 19.30          & 23.97            & 26.79           & 24.11 \\ \cline{2-8}
          & 11.10     & 20.88    & 20.98      & 13.13          & 21.09            & 22.53           & 21.31 \\ \hline
          
\multirow{4}{*}{\includegraphics[width=0.06\linewidth]{boat_orig}}
          & 28.16     & 32.05    & 32.46      & 27.63          & 32.38            & 32.93           & 33.13 \\ \cline{2-8}
          & 22.24     & 29.48    & 29.20      & 26.23          & 29.07            & 30.44           & 29.34 \\ \cline{2-8}
          & 15.06     & 25.35    & 25.12      & 21.64          & 25.10            & 26.38           & 24.59 \\ \cline{2-8}
          & 11.01     & 22.05    & 22.14      & 13.42          & 22.20            & 23.11           & 22.69 \\ \hline
          
\multirow{4}{*}{\includegraphics[width=0.06\linewidth]{cameraman_orig}}                                                                                                                     
          & 28.34     & 31.82    & 32.62      & 26.77          & 32.51            & 32.40           & 32.79 \\ \cline{2-8}
          & 22.50     & 28.37    & 29.05      & 24.41          & 28.48            & 29.68           & 28.60 \\ \cline{2-8}
          & 15.30     & 23.88    & 24.04      & 20.37          & 23.68            & 25.14           & 23.35  \\ \cline{2-8}
          & 11.13     & 19.98    & 20.29      & 13.84          & 20.18            & 21.23           & 20.54 \\ \hline
          
\end{tabular}
\label{compare2}
\end{table*}

\begin{figure}[!t]
  \centering
  \includegraphics[width=0.85\linewidth]{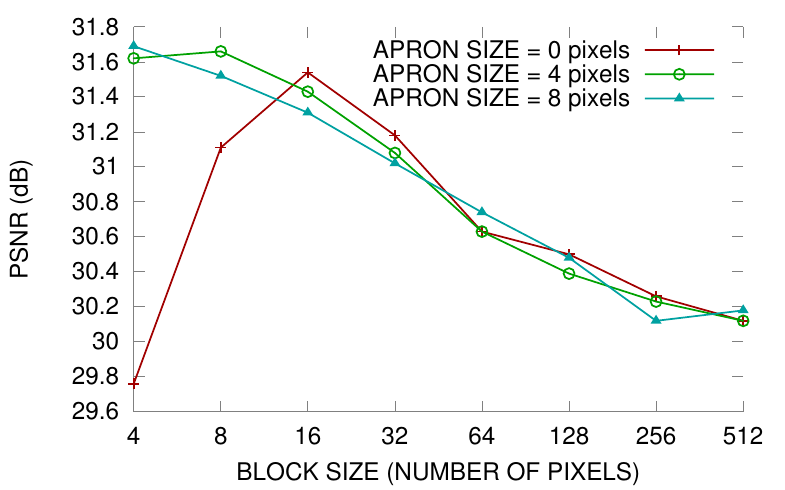}
  \caption{Variation of PSNR with block size for different apron sizes. This result is for the 512$\times$512 8-bit \textit{Lenna} image with a Gaussian filter size of 9$\times$9 pixels.}
  \label{psnrplot}
\end{figure}

\begin{figure}[!t]
  \centering
  \subfigure{\includegraphics[width=0.6\linewidth]{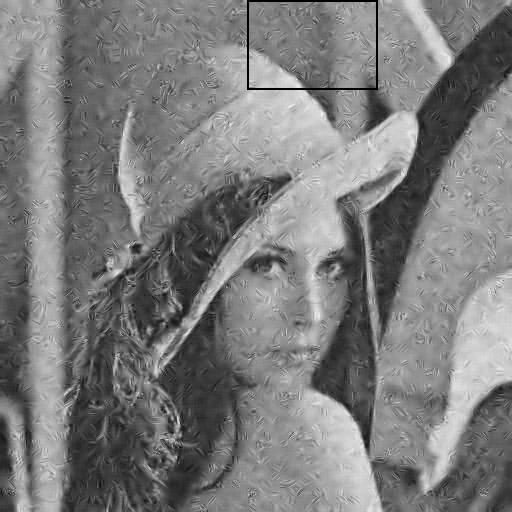}}
  \subfigure{\includegraphics[width=0.3\linewidth]{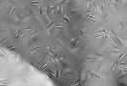}}
  \caption{Freckles in the denoised image when no apron is used.  The block size is 8 $\times$ 8, and the input PSNR is 14.61~dB.}
  \label{freckles}
\end{figure}

\begin{figure}[!t]
  \centering  
  \subfigure{\includegraphics[width=0.6\linewidth]{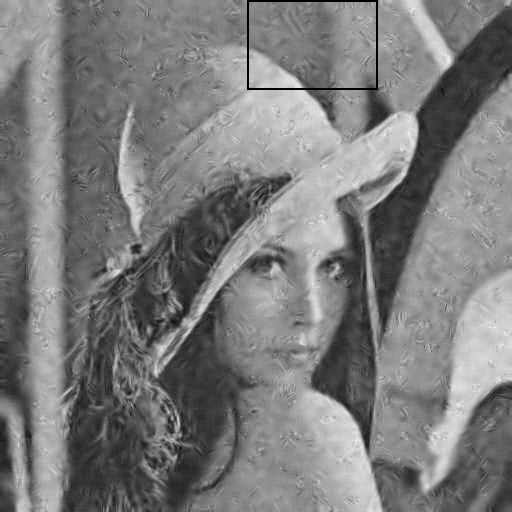}}
  \subfigure{\includegraphics[width=0.3\linewidth]{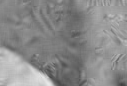}}
  \caption{Less freckles in the denoised image when a 4~px apron is used.  The block size is 8 $\times$ 8 pixels, and the input PSNR is 14.61~dB.}
  \label{less_freckles}
\end{figure}

\indent The proposed spatially varying Gaussian smoother (SVGS) was prototyped\footnote{The prototype image denoising program is available under the MIT open-source license at \url{https://github.com/s-gv/image-denoise-parallel}.} in the C programming language. OpenCL was used for parts of the program accelerated by the Graphics Processing Unit (GPU). All experiments were conducted on an Intel Core i3-2100 CPU running at 3.10~GHz with 8 GB of main memory. The GPU used was an AMD Radeon HD 7950 running at 850~MHz with 3~GB of GDDR5 RAM. Linux operating system with kernel version 3.8.0-29-generic x86\_64 was used. The conjugate gradient optimizer was used to minimize the Stein-free unbiased risk estimate of the MSE.\\
\indent Figure~\ref{psnrplot} shows the denoising performance of the proposed algorithm for different block and apron sizes. The general trend is that the PSNR decreases as the block size is increased. However, for small block sizes, the PSNR falls because the cost function estimate becomes unreliable. Including an ``apron" of additional pixels around each block, when computing the cost function, alleviates the problem to a certain extent. Also, for smaller block sizes with no apron, the filter may excessively tune to the local orientation leading to freckles in the denoised image (cf. Figure~\ref{freckles}). A non-zero apron was found to give rise to less freckles (cf. Figure~\ref{less_freckles}).\\
\begin{figure}[!t]
  \centering
  \includegraphics[width=0.85\linewidth]{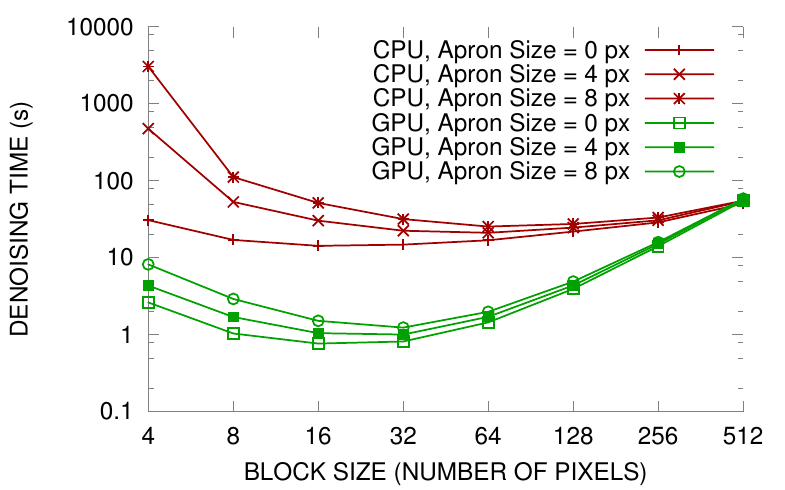}
  \caption{A log-log plot of denoising time against block size for different apron size. This figure is for the 512$\times$512 8-bit \textit{Lenna} image with a filter size of 9$\times$9 pixels.}
  \label{runtime}
\end{figure}
\indent Figure~\ref{runtime} shows the amount of time required to denoise an image for different block and apron sizes. The CPU version runs the entire denoising algorithm on the CPU, whereas the GPU version offloads filter creation and cost function computation to the GPU. For small block sizes, the GPU-accelerated version is about 10$\times$ to 400$\times$ faster than the CPU-only version. As the block size increases, the computation time decreases since the absolute number of apron pixels decreases with increasing block size. However, after a certain point, increasing the block size results in an increase in the compute time. We believe that this is largely an artifact of our implementation, which takes advantage of pixel-level parallelism only when performing convolution and not for computing the summation necessary to find the cost function.\\
\indent For the 512$\times$512 8-bit \textit{Lenna} image with input PSNR 22.12~dB, a block size of 4$\times$4 pixels with an apron size of 8$\times$8 pixels gave the highest output PSNR of 31.69~dB in 8.2 seconds with GPU acceleration. However, using a block size of 8$\times$8 pixels with an apron size of 4$\times$4 pixels gave a PSNR of 31.66 dB in only 1.7 seconds. Thus, we chose the block size to be 8$\times$8 pixels and apron size to be 4$\times$4 pixels.\\
\indent In Tables~\ref{compare} and \ref{compare2}, we show a performance comparison of the proposed algorithm with some state-of-the-art denoising techniques. The proposed technique performs reasonably well for both Gaussian and Laplacian noise and falls short of BM3D\cite{bm3d} by about 1~dB. The performance is on par with orthonormal wavelet thresholding (OWT), ProbShrink, BiShrink, Multiresolution bilateral filter for most input PSNRs. In the case of the OWT, which relies on risk estimation, the output PSNR dips by about 3 dB when the noise statistics change from Gaussian to Laplacian although the input PSNR remains the same. The decrease in performance is significant at lower input PSNRs. It must be emphasized that the state-of-the-art techniques are much more sophisticated than the proposed method. For the \textit{Barbara} image, the trend in performance is different, and the output PSNR is slightly poorer than that of the other techniques (compared with that for the other images). The reason is our choice of the Gaussian filter for denoising, which is not optimal for smoothing texture. We chose the Gaussian smoother mainly for illustration and one could think of other linear filters for denoising. For example, one could use the Gabor kernel for smoothing texture patterns. There is also physiological evidence that the cells in the mammalian visual cortex work in the spatial domain in a manner similar to Gabor filters \cite{neuro_jones1987evaluation}.

\begin{figure}
  \centering
  \subfigure{\includegraphics[width=0.6\linewidth]{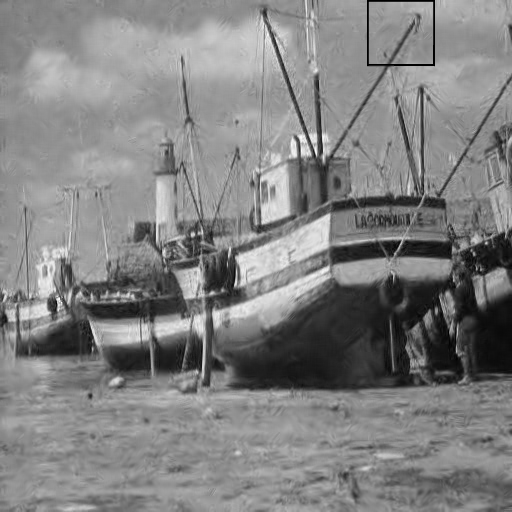}}
  \subfigure{\includegraphics[width=0.3\linewidth]{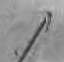}}
  \caption{Denoised \textit{Boats} image. Input PSNR is 22.17~dB, and output PSNR is 29.74~dB.}
  \label{boat_denoised}
\end{figure}

\section{Conclusions}
\label{conc}
\indent We addressed the problem of risk estimation for image denoising, where the goal mainly is to arrive at a reliable and accurate substitute for the mean-square error. In most image processing literature, one often assumes a particular noise distribution and then develops risk estimators using Stein's lemma (for Gaussian noise) and its counterparts for other noise types (for example, the Hudson's identity for Poisson noise \cite{hudson}). We showed that in the additive white noise scenario, assuming a linear denoising function, one does not need the Stein lemma or its counterpart to arrive at a risk estimator. The advantage is that the risk estimator becomes agnostic to the distribution of noise, which may be a good thing to have particularly in scenarios where such information is not available or cannot be estimated reliably. The downside is that the proposed approach is applicable only for linear denoising functions, whereas the Stein-type approaches are suitable even for nonlinear denoising functions. Notwithstanding the drawback, we showed that the denoising performance obtained with a simple Gaussian smoother with its variance parameter optimized on a patch-by-patch basis gave denoising performance that is competitive with far more sophisticated state-of-the-art denoising techniques, for both Gaussian and Laplacian noise contamination. The parallelism in the proposed method also made the implementation amenable to GPU-based acceleration. By considering more sophisticated linear filters, one could further improve upon the overall denoising performance at the cost of increased optimization overhead.

\section*{Acknowledgments}
\indent We would like to thank Harini Kishan and Subhadip Mukherjee for technical discussions.

\bibliographystyle{IEEEtran}

\end{document}